\documentclass[10pt,twocolumn,letterpaper]{article}

\usepackage{cvpr}              

\usepackage{graphicx}
\usepackage{amsmath}
\usepackage{amssymb}
\usepackage{booktabs}
\usepackage{float}
\usepackage{multirow}
\usepackage{array}
\usepackage[accsupp]{axessibility}

\usepackage[pagebackref,breaklinks,colorlinks]{hyperref}

\usepackage[capitalize]{cleveref}
\crefname{section}{Sec.}{Secs.}
\Crefname{section}{Section}{Sections}
\Crefname{table}{Table}{Tables}
\crefname{table}{Tab.}{Tabs.}

\def\cvprPaperID{7057} 
\def\confName{CVPR}
\def\confYear{2023}

\begin{document}

\title{Continuous Intermediate Token Learning with Implicit Motion Manifold for Keyframe Based Motion Interpolation}

\author{Clinton A. Mo\textsuperscript{1},
Kun Hu\textsuperscript{1,}\thanks{Corresponding author.} ,
Chengjiang Long\textsuperscript{2}, 
Zhiyong Wang\textsuperscript{1}\\
\textsuperscript{1}{School of Computer Science, The University of Sydney, NSW 2006, Australia} \\
\textsuperscript{2}{Meta Reality Labs, Burlingame, CA, USA} \\
{\tt\small clmo6615@uni.sydney.edu.au, \{kun.hu, zhiyong.wang\}@sydney.edu.au, clong1@meta.com}
}

\maketitle
\begin{abstract}
Deriving sophisticated 3D motions from sparse keyframes is a particularly challenging problem, due to continuity and exceptionally skeletal precision. The action features are often derivable accurately from the full series of keyframes, and thus, leveraging the global context with transformers has been a promising data-driven embedding approach. However, existing methods are often with inputs of interpolated intermediate frame for continuity using basic interpolation methods with keyframes, which result in a trivial local minimum during training. In this paper, we propose a novel framework to formulate latent motion manifolds with keyframe-based constraints, from which the continuous nature of intermediate token representations is considered. Particularly, our proposed framework consists of two stages for identifying a latent motion subspace, {\em i.e.}, a keyframe encoding stage and an intermediate token generation stage, and a subsequent motion synthesis stage to extrapolate and compose motion data from manifolds. Through our extensive experiments conducted on both the LaFAN1 and CMU Mocap datasets, our proposed method demonstrates both superior interpolation accuracy and high visual similarity to ground truth motions. {\let\thefootnote\relax\footnote{Code available at: \url{https://github.com/MiniEval/CITL}}}
\end{abstract}

\section{Introduction}
\label{sec:intro}

Pose-to-pose keyframing is a fundamental principle of character animation, and animation processes often rely on key pose definitions to efficiently construct motions \cite{frank1981disney, Dang:ICCV2021, mo2021keyframe}. In computer animation, keyframes are temporally connected via interpolation algorithms, which derive intermediate pose attributes to produce smooth transitions between key poses. However, human motion is often complex and difficult to be effectively represented by sparse keyframe sequences alone. While this can be addressed by producing denser sequences of key poses, this approach is laborious for animators, thereby increasing the cost of keyframed animation processes. Even with Motion Capture (MoCap) workflows, artists must often resort to keyframing in order to clean artifacts, impose motion constraints, or introduce motion features irreplicable by motion capture performers. 

 \begin{figure}[t]
     \centering
     \includegraphics[width=\linewidth]{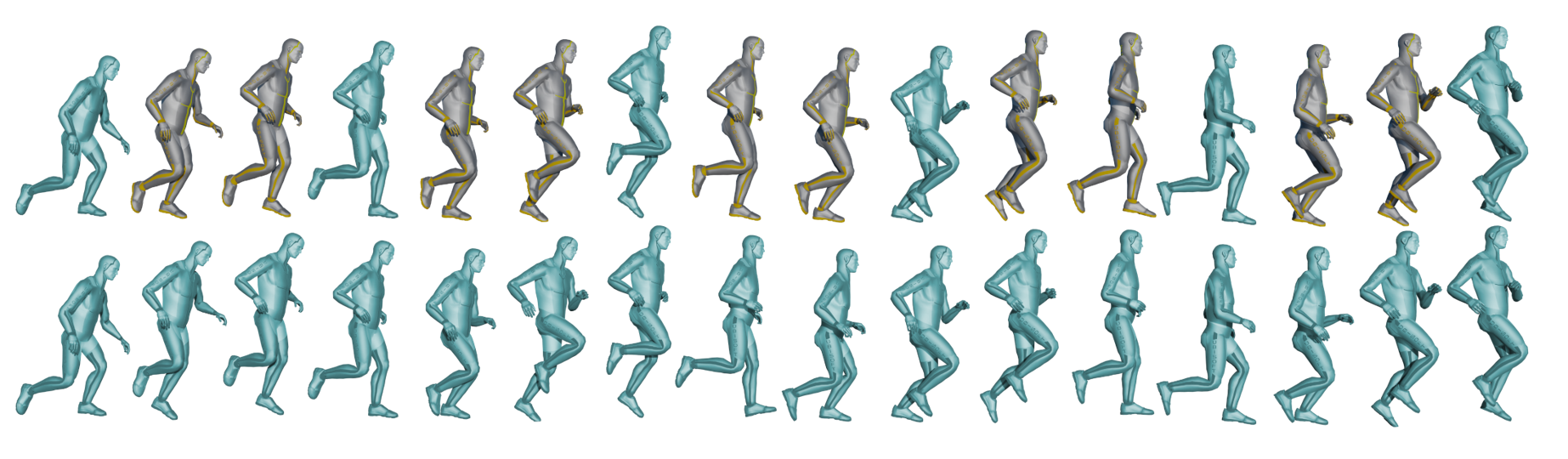}
     \caption{An example of motion interpolation by our method (first row), given the keyframes of a hopping motion (in blue), compared with the ground truth (second row). 
     }
     \label{fig:hop}
 \end{figure}

Learning-based motion interpolation methods have recently been proposed as an acceleration of the keyframed animation process, by automatically deriving details within keyframe transitions as shown in Figure \ref{fig:hop}. Various machine learning methods have been explored to enable more realistic interpolation solutions from high quality MoCap databases, e.g. by using recurrent networks \cite{harvey2018recurrent, harvey2020robust, tang2022real} or transformer-based approaches \cite{duan2021single,oreshkin2022motion,liu2021motion}. 
Guiding data-driven interpolation with real motions is particularly attractive for keyframe animation workflows, as realistic motions often require the greatest amount of keyframing, by virtue of their subtle motion details and physical constraints. 

Naturally, as a sequence in-painting problem, motion interpolation can be formulated as a masked sequence-to-sequence task, which the recent popular transformer approach is expected to learn effectively \cite{song2019mass, vaswani2017attention, collrepresenting,wang2023multi}. However, sequential learning of masked continuous attributes with transformers is largely impaired by the conventional masked tokens for intermediate data. A token is defined as an individual data element on the extendable axis of a sequence, namely the temporal axis for motions. In current sequence modelling formulations, a token is usually represented by a one-hot vocabulary vector to specify individual words or masked elements, which poses a limitation on continuous attributes.
Since continuous attributes can be assigned any real value, there exists no value by which a masking token can be defined without corresponding to an otherwise valid input. Previous approaches have employed transformer decoder-level mask tokens and linear interpolation (LERP)-based tokens have been explored to work around this issue \cite{he2022masked, duan2021single, oreshkin2022motion}.
However, these approaches have innate incompatibilities with the transformer architecture. Singular mask token representations, regardless of their point of introduction, result in discontinuous hidden representations, which are antithetical to the evaluation of continuous motion data. On the other hand, the use of LERP as a pre- or post-processing step necessarily introduces an accurate starting estimate to the solution, which transformer models are prone to becoming over-reliant on \cite{xiong2020layer, liu2020understanding}.
To fully address these limitations, we propose a novel transformer-based framework that learns to model keyframe sequences into latent motion manifold representations for intermediate tokens, which reflects the smooth and continuous nature of human motion.

As illustrated in Figure~\ref{fig:arch}, our proposed framework incorporates three stages with transformers to convert a keyframe sequence into a complete motion sequence: 
Stage-I is a \textit{keyframe encoding stage} to formulate the overall motion patterns from the keyframe sequence into keyframe context tokens as a guidance for further modelling; 
Stage-II is an \textit{intermediate token generation stage}, where temporal indices are mapped into intermediate token representations with the keyframe context tokens, which serve as an implicit latent motion manifold constraint; and 
Stage-III, a \textit{motion synthesis stage}, takes the obtained intermediate tokens by injecting them within the keyframe token sequence, and interpolating them to derive a refined motion sequence estimation. 

With this framework, our transformer-based approach exhibits two key advantages over existing approaches that enable its high-quality motion interpolation: a) Manifold learning allows our framework to establish temporal continuity in its latent representation space, and b) The latent motion manifold constrains our transformer model to concentrate its attention exclusively towards motion keyframes, as opposed to intermediate tokens derived from non-keyframe poses, such as those derived from LERP, thereby forcing a necessary alignment between the known and unknown tokens adaptively.


In addition, we identify an adverse link between continuous features and normalisation methods with per-token re-centering. Specifically, layer normalisation (LayerNorm) \cite{ba2016layer}, which is commonly used in transformer architectures, constrains the biases of token features based on their individual distributions. Though this is well-known to be effective with linguistic models \cite{vaswani2017attention, liu2019roberta}, continuous data inherently contain biases that should be leveraged at sequence level. Therefore, we introduce a sequence-level re-centering (Seq-RC) technique, where positional pose attributes of keyframes are recentred based on their distribution throughout a motion sequence, and root-mean-square normalisation (RMSNorm) \cite{zhang2019root} layers are then employed to perform magnitude-only normalisation. Though RMSNorm was initially proposed as only a speedup to LayerNorm, our observations demonstrate that 
Seq-RC leads to superior performance in terms of accuracy and visual similarity to MoCap sequences.

In summary, our paper's key contributions are threefold:

\begin{enumerate}
    \item We propose a novel transformer-based architecture consisting of three cooperative stages. It constrains the evaluation of unknown intermediate representations of continuous attributes to the guidance of keyframe context tokens in a learned latent manifold. 
    \item We devise sequence-level re-centering (Seq-RC) normalisation to effectively operate with real scalar attributes with minimal accuracy loss. 
    \item Extensive comparisons and ablation results obtained on LaFAN1 and CMU Mocap strongly demonstrate the superiority of our method over the state-of-the-art.
\end{enumerate}

\section{Related work}

In this section, we explore existing machine learned methods by which motion keyframes, or key tokens of other mediums, can be used to generate full sequences. We also review known methods for intermediate data prediction.

\subsection{Motion synthesis and completion}

The necessities of motion interpolation techniques have existed since the early days of computer animation. A key advantage of computer animation over traditionally drawn animation is its ability to automatically evaluate a smooth motion from a keyframe representation. The widely accepted method for applying interpolation to motion keyframes is through the use of function curves (F-Curves), by which various mathematical functions can be defined between intervals. The common functions used today for keyframed motion representation are often based on Bézier spline curves \cite{snibbe1995direct,jafari2014spherical}, though any function can technically be employed for this purpose. In addition, inverse kinematics-based constraints \cite{rose1996efficient} have been used jointly with Markov chain methods \cite{lehrmann2014efficient} and decision graphs \cite{kovar2008motion} to generate constrained keyframe or motion sequence transitions in interactive applications such as 3D video games. 

More recently, deep learning methods have enabled effective motion completion from sparse keyframe representations. By formulating motion interpolation as a motion synthesis problem with keyframe constraints, a neural network-assisted keyframed animation approach is emerging as a more effective solution to the current approaches. Recurrent neural networks have been able to derive realistic motion details from motion keyframe compositions \cite{zhang2018data, kundu2019bihmp, harvey2020robust} and real-time control schemes \cite{tang2022real}. In addition, sequence masking approaches such as BERT-based and autoencoder-based methods \cite{kaufmann2020convolutional,duan2021single,liu2021motion,oreshkin2022motion, Ma:CVPR2022, dang2022diverse} have enabled full keyframe sequence analysis for completing motions with a more comprehensive context. However, these methods are severely affected by the tendency of transformer-based networks that toward a trivial local minimum, given an initial LERP starting point. This limitation in transformers has been thoroughly observed and documented as the result of early gradient instabilities in attention weights \cite{vaswani2017attention, xiong2020layer, liu2020understanding, bai2021transformers}. 

Learned pose manifolds have become a prominent approach to synthesis plausible human poses, which restrain pose attributes to a specified space \cite{tiwari2022pose, pavlakos2019expressive}. These methods mainly focus on dense poses from complete motion sequences. Our work extends this concept for an incomplete and sparse scenario, by using the keyframes as constraints to derive an implicit motion manifold. 

\begin{figure*}
    \centering
    \includegraphics[width=\linewidth]{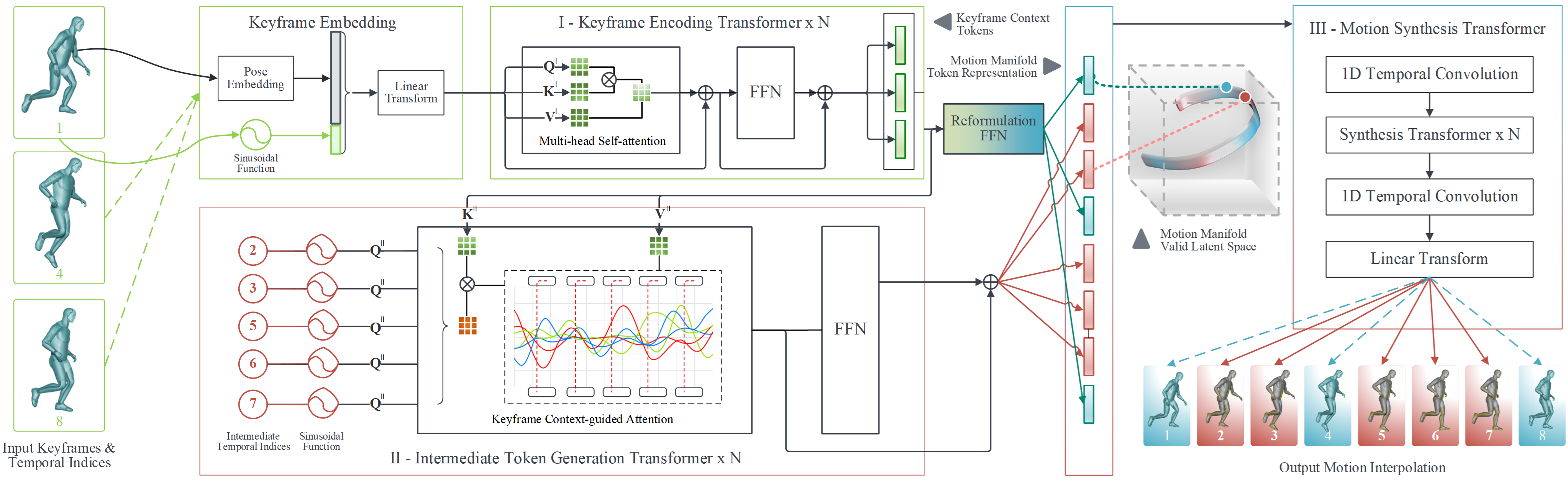}
    \caption{Overview of our transformer architecture and its three main components: (I) Encoding transformer $\Phi^{\text{key}}$, (II) Intermediate token generation transformer $\Phi^{\text{imd}}$, and (III) Motion synthesis transformer $\Phi^{\text{syn}}$.}
    \label{fig:arch}
\end{figure*}

\subsection{Transformer-based temporal in-painting}

Inter-frame video interpolation is a similar task to motion interpolation, due to its common goal of predicting transitions between frames on the temporal axis. Like motion in-painting methods, temporal transformers for video interpolation use blended inputs as masks for continuous attributes \cite{lu2022video}. Alternatively, when interpolating with an individual interval, the mask tokens can be copied from the previous keyframe, with positional encoding being the sole difference between input tokens \cite{liu2020convtransformer}. While the feature extraction process of video data can take full advantage of the global context afforded by using transformers, other mechanisms such as convolutions can be integrated with a transformer for video in-painting as well \cite{shi2022video}. For skeleton-based motion data, existing studies suggest that graph convolutional networks can improve analytic and synthetic performance \cite{mo2021keyframe,liu2021motion,guo2019human,degardin2022generative}; however, the application of pure convolutional approaches for data synthesis is only valid when the interval length between keyframes is constant. 

\subsection{Masked data modelling}

Various mask-based machine learning techniques have been proposed to estimate missing values of incomplete sequences based on their known values. The use of masked token is well known to be highly effective for producing pre-trained linguistic transformer models \cite{vaswani2017attention,dong2019unified,liu2019roberta,song2019mass}. However, due to aforementioned limitations on token representations for continuous attributes, adapting transformer-based masked data modelling for computer vision tasks has largely focused on masking schemes. Discretised tokens for visual data is a proposed workaround \cite{wang2022bevt, bao2021beit}; however, the information loss of tokenisation renders this technique unfeasible for precise mediums like motion data. Masked auto-encoders assign mask tokens to an encoded latent space \cite{he2022masked, wang2022bevt}, and adopt them to masked sequences at the decoder level. This allows the transformer encoder to learn solely from known tokens; however, the monolithic definition of masked tokens results in a discontinuous sequence representation in latent space.  

Additionally, masked convolutional networks have been a consistently effective approach for image-based data. The popular U-Net structure of convolutional models \cite{ronneberger2015u} involves a data bottleneck, which elicits behaviours for ignoring masked image regions \cite{iizuka2017globally,song2019geometry,yang2020learning}. 
However, the pooling technique of convolutional models restrains the learned features to a fixed scale, and struggles to extrapolate global features with increased data scales, e.g. higher image resolutions or longer videos. Conversely, transformer-based methods are particularly effective with long-range feature learning, and as such, we believe them to be the more suitable for keyframe-guided sequential motion learning.

\section{Methodology}

We formulate the motion interpolation task as a data imputation problem from a sparse representation. The primary objective of our solution is to define motion sequence representations as a function of motion keyframes, which we learn using a transformer-based model. 
As shown in Figure \ref{fig:arch}, our solution comprises of three cooperative stages to learn a latent motion manifold using transformers to convert a sequence $K = \{x_0, x_{t_1}, x_{t_2}, ..., x_{N-1}\}$ at frames $T_K = \{0, t_1, t_2, ..., N-1\}$ into a complete motion sequence $Y = \{y_0, y_1, y_2, ..., y_{N-1}\}$ with temporal indices $T = \{0,1,2,...,N-1\}$ (i.e. $T_K \subset T$):

\begin{itemize}
    \item Stage-I, a \textbf{keyframe encoding stage} formulates the keyframe sequence $K$ into keyframe context tokens $\Phi^{\text{key}}(K) = \{\phi_0, \phi_{t_1}, \phi_{t_2}, ..., \phi_{N-1}\}$, which serves as the motion encoding for further modelling.
    \item Stage-II, an \textbf{intermediate token generation stage} maps the temporal indices $t\in T \setminus T_K$ into intermediate tokens $\Phi^{\text{imd}}(t|\Phi^{\text{key}}(K))$ with the guidance of keyframe context, which constrains the latent space to obtain an implicit motion manifold. 
    \item Stage-III, a \textbf{motion synthesis stage} injects the intermediate tokens between the keyframe tokens $\Phi^{\text{key}}(K)$, and decodes the resulting token sequence with a transformer $\Phi^{\text{syn}}$ to derive a motion sequence $\hat{Y} = \{\hat{y}_0, \hat{y}_1, ..., \hat{y}_{N-1}\}$ as an estimation of $Y$.
\end{itemize}

\subsection{Keyframe poses to motion sequence context}

Our pose representations are comprised of seven elements per joint: $P_t \in \mathbb{R}^{J \times 3}$ values for global 3D positions produced through forward kinematics (FK), and $q_t \in \mathbb{R}^{J \times 4}$ values for unit quaternion representations of local rotations, that is, $x_t = [P_t, q_t] \in \mathbb{R}^{J \times 7}$, where $J$ represents the number of joints. In motion capture, local 3D positions for each joint are generally constant, determined by its positional offsets from its parent joint, and do not require explicit representation for the input. The sole exception to this is the root joint, where the global position is the local position. 

In Stage-I, we encode the keyframe sequence $K$ into a learned keyframe context token representation $\Phi^{\text{key}}(K)$. It is used as a feature map for both the intermediate token generation and motion synthesis stages. First, we project the pose data $x_t$, $t\in T_K$, of each keyframe into a pose embedding vector $x^\prime_t\in\mathbb{R}^{d}$ using a linear layer, where $d$ is the token embedding dimension. Next, we adopt sinusoidal positional encoding (PE) $PE_{pos}$ \cite{vaswani2017attention} of $n$-dimension to $x^\prime_t$.
Unlike natural language processing (NLP) approaches, we do not add PE to our token representations, but instead concatenate a fixed-length PE for two key reasons:
\begin{itemize}
    \item PE acts as a sliding binary vector, and thus can represent $2^n$ positions using $n$ elements. For our task, we set $n$ to 16, allowing our PE vector to produce $2^{16}$ positions, which is sufficient for our purposes.
    \item Additive PE introduces minor disruptions in token representations. While the discrete data of NLP can benefit from slight variations in token representations, it acts as a hindrance in continuous data due to the necessity of precision. 
\end{itemize}

\subsection{Sequence-level re-centering normalisation}

Our Stage-I transformer $\Phi^{\text{key}}$ includes a number of layers based on multi-head self-attentions, each followed by GELU-activated feed-forward networks (FFNs). 
We replace all instances of LayerNorm the transformer encoders with RMSNorm \cite{zhang2019root}, which does not involve a re-centering step. 
This avoids the token-level re-centering present in LayerNorm, which we observed to be detrimental for our regression tasks. We believe that this is due to feature biases that are present in and crucial to continuous attributes, e.g. a pose with a high root position values will result in all joints having similarly high position values.
Therefore, we introduce a sequence level re-centering scheme for normalisation at the transformer input-level only
, re-centreing the root positions of the input based on the mean root positions of all keyframes in the input sequence. 

\subsection{Motion manifold with context-guided attention}

In Stage-II, our intermediate token generation transformer $\Phi^{\text{imd}}$ aims to learn motion manifolds navigable by temporal indices $t\in\mathbb{N}$ by the guidance of the keyframing context from $\Phi^{\text{key}}(K)$. 
In detail, we accomplish this through a keyframe context-guided attention mechanism, where the attention derives key and value mappings exclusively from linear transformations of $\Phi^{\text{key}}(K)$. For each intermediate token, the query is simply a sinusoidal embedding of the token's temporal position. 

We constrain our manifold implicitly using two mechanisms. Firstly, the intermediate tokens are entirely sourced as a product of $\Phi^{\text{key}}(K)$ value transformations, which inherently limits the range of latent representations in the manifold. Secondly, the 1D convolutional layers in Stage-III entail feature dependencies between temporally adjacent tokens, which enables the disparately obtained $\Phi^{\text{key}}$ and $\Phi^{\text{imd}}$ tokens to converge towards coordination. 

\subsection{Interpolated motion synthesis}

In Stage-III, a transformer $\Phi^{\text{syn}}$ is introduced to take the intermediate token representations $m_t$ and estimate the complete motion sequence $\hat{Y}$. 
Since the primary role of keyframe context tokens $\Phi^{\text{key}}(K)$ is to derive the intermediate token embeddings, an additional projection is performed using a single FFN $\texttt{FFN}(\Phi^{\text{key}}(K))$ to reformulate the keyframing context tokens into keyframe tokens that adhere to the motion manifold. 
The intermediate tokens obtained from $\Phi^{\text{imd}}$ are ready to be used directly. To this end, we can construct the resulting motion manifold $\hat{M}$ with a token sequence $\{\hat{m}_0, \hat{m}_1, ..., \hat{m}_{N-1}\}$ as follows:
\begin{equation}
\small
\label{equ:ff_m_t}
    \hat{m}_t = \begin{cases}
        \texttt{FFN}(\Phi^{\text{key}}(K)_t), & \text{if } t \in K\\
        m_t, & \text{otherwise}
    \end{cases}
\end{equation}
where $\Phi^{\text{key}}(K)_t$ is the context token representation for the keyframe of index $t$ in the sequence. 

Before applying the transformer $\Phi^{\text{syn}}$, we feed the tokens of $\hat{M}$ through a 1D convolutional layer of kernel size 3. For each layer of $\Phi^{\text{syn}}$, a self-attention function is followed by a FFN. The output token sequence of $\Phi^{\text{syn}}$ is fed through another 1D convolutional layer before the final linear projection. The final output $\hat{Y}_t$ consists of a root position estimation $\hat{p}_{t,0} \in \mathbb{R}^3$ of $p_{t,0}$ and local quaternions $\hat{q}_t \in \mathbb{R}^{J \times 4}$. Note that $p_{t,0}$ and $q_t$ jointly can be used to compute the global position $P_t$ and rotation $Q_t$ information by FK. 

\subsection{Loss functions}

The stages in our method are trained jointly in an end-to-end manner using a set of loss functions. To determine the loss for a motion sequence of length $N$, we use $\ell_1$ distance for the following features obtained from $\hat{Y}$ and $Y$:

\begin{itemize}
    \item \textbf{Local/root position loss}: With predicted and real coordinate values of the root joint at the $t$-th frame as $\hat{p}_{t,0} \in \mathbb{R}^3$ and $p_{t,0} \in \mathbb{R}^3$ respectively,
    \begin{equation}
    \footnotesize
        L_{root} = \frac{1}{N} \sum_{t=0}^{N-1} ||\hat{p}_{t,0} - p_{t,0}||_1.
    \end{equation}
    \item \textbf{Local rotation loss}: With predicted (pre-normalised) and expected (unit-normalised) quaternion values of all joints at frame $t$ as $\hat{q}_{t,j} \in \mathbb{R}^{4}$ and $q_{t,j} \in \mathbb{R}^{4}$ respectively,
    \begin{equation}
    \footnotesize
        L_{quat} = \frac{1}{NJ} \sum_{t=0}^{N-1}\sum_{j=0}^{J-1} ||\hat{q}_{t,j} - q_{t,j}||_1.
    \end{equation}
    \item \textbf{Global position loss}: With predicted and real FK-derived coordinate values of all joints at frame $t$ as $\hat{P}_t \in \mathbb{R}^{J \times 3}$ and $P_t \in \mathbb{R}^{J \times 3}$ respectively,
    \begin{equation}
    \footnotesize
        L_{FK_p} = \frac{1}{NJ} \sum_{t=0}^{N-1} ||\hat{P}_t - P_t||_1.
    \end{equation}
    \item \textbf{Global rotation loss}: With predicted and real FK-derived quaternion values of all joints at frame $t$ as $\hat{Q}_t \in \mathbb{R}^{J \times 4}$ and $Q_t \in \mathbb{R}^{J \times 4}$ respectively,
    \begin{equation}
    \footnotesize
        L_{FK_q} = \frac{1}{NJ} \sum_{t=0}^{N-1} ||\hat{Q}_t - Q_t||_1.
    \end{equation}
\end{itemize}

Although quaternions are unit-normalised in practice, we found that calculating $L_{quat}$ with non-normalised predictions resulted in improved gradient stability during the training process and, in turn, greater training convergence.

In summary, our training loss function $L$ is as follows:
\begin{equation}
    \small
    L = \alpha_l (L_{root} + L_{quat}) + \alpha_g (L_{FK_p} + L_{FK_q}),
\end{equation}
where $\alpha_l$ and $\alpha_g$ are local and global feature loss scaling parameters, respectively. The accuracy of local attributes is best prioritised over that of global attributes, since normalised quaternions remain in use for deriving global features, which lead to gradient instability \cite{harvey2020robust, duan2021single}.

\section{Experiments and results}

\begin{table*}
    \centering
    \tiny
    \setlength\tabcolsep{1.2pt}
    \resizebox{\textwidth}{!}{
    \begin{tabular}{|c|c c c|c c c|c c c|c c c|c c c|c c c|c c c|c c c|c c c|}
        \hline
        Category & \multicolumn{9}{c|}{CMU:acrobatics} & \multicolumn{9}{c|}{CMU:basketball} & \multicolumn{9}{c|}{CMU:golf} \\
        \hline
         & \multicolumn{3}{c|}{\textbf{L2P}} & \multicolumn{3}{c|}{\textbf{L2Q}} & \multicolumn{3}{c|}{\textbf{NPSS}} & \multicolumn{3}{c|}{\textbf{L2P}} & \multicolumn{3}{c|}{\textbf{L2Q}} & \multicolumn{3}{c|}{\textbf{NPSS}} & \multicolumn{3}{c|}{\textbf{L2P}} & \multicolumn{3}{c|}{\textbf{L2Q}} & \multicolumn{3}{c|}{\textbf{NPSS}}\\
        Interval & 5 & 15 & 30 & 5 & 15 & 30 & 5 & 15 & 30 & 5 & 15 & 30 & 5 & 15 & 30 & 5 & 15 & 30 & 5 & 15 & 30 & 5 & 15 & 30 & 5 & 15 & 30 \\
        \hline
        LERP & 0.234 & 0.899 & 1.718 & 0.308 & 1.052 & 1.598 & 0.183 & 0.854 & 1.856 & 0.122 & 0.544 & 1.051 & 0.182 & 0.695 & 1.044 & 0.078 & 0.352 & 1.078 & \textbf{0.029} & 0.181 & 0.509 & \textbf{0.044} & 0.196 & 0.473 & \textbf{0.060} & 0.396 & 1.296 \\
        \hline
        BERT & 0.353 & 0.947 & 1.744 & 0.516 & 1.211 & 1.712 & 0.249 & 0.886 & 1.865 & 0.165 & 0.547 & 1.043 & 0.221 & 0.734 & 1.075 & 0.070 & 0.296 & 0.972 & 0.066 & 0.181 & 0.511 & 0.071 & 0.197 & 0.475 & 0.078 & 0.280 & 1.187 \\
        $\Delta$-interpolator & \textbf{0.189} & 0.727 & 1.395 & \textbf{0.270} & 0.814 & 1.342 & \textbf{0.152} & 0.642 & 1.560 & 0.140 & 0.594 & 1.212 & 0.206 & 0.717 & 1.089 & 0.089 & 0.344 & 1.023 & 0.056 & 0.193 & 0.511 & 0.085 & 0.216 & 0.487 & 0.078 & 0.390 & 1.276 \\
        $\text{TG}_{\text{complete}}$ & 0.400 & 0.982 & 1.624 & 0.499 & 1.177 & 1.772 & 0.297 & 1.004 & 2.054 & 0.181 & 0.482 & 0.921 & 0.255 & 0.718 & 1.116 & 0.106 & 0.291 & 0.769 & 0.113 & 0.222 & 0.461 & 0.121 & 0.244 & 0.458 & 0.110 & 0.208 & 0.635 \\
        MAE & 0.295 & 0.697 & 1.156 & 0.730 & 1.219 & 1.511 & 0.444 & 0.878 & 1.444 & 0.163 & 0.405 & 0.739 & 0.252 & 0.620 & 0.915 & 0.095 & 0.255 & 0.855 & 0.080 & 0.139 & 0.214 & 0.110 & 0.185 & 0.267 & 0.114 & 0.177 & 0.316 \\
        \hline
        Ours & 0.217 & \textbf{0.471} & \textbf{0.791} & 0.328 & \textbf{0.643} & \textbf{0.931} & 0.167 & \textbf{0.410} & \textbf{0.769} & \textbf{0.101} & \textbf{0.274} & \textbf{0.456} & \textbf{0.165} & \textbf{0.473} & \textbf{0.668} & \textbf{0.054} & \textbf{0.160} & \textbf{0.329} & 0.048 & \textbf{0.083} & \textbf{0.116} & 0.083 & \textbf{0.138} & \textbf{0.182} & 0.081 & \textbf{0.104} & \textbf{0.155} \\
        \hline
    \end{tabular}
    }
    \resizebox{\textwidth}{!}{
    \begin{tabular}{|c|c c c|c c c|c c c|c c c|c c c|c c c|c c c|c c c|c c c|}
        \hline
        Category & \multicolumn{9}{c|}{CMU:salsa} & \multicolumn{9}{c|}{CMU:swim} & \multicolumn{9}{c|}{CMU:walk/run} \\
        \hline
        LERP & 0.272 & 1.091 & 2.087 & \textbf{0.362} & 1.265 & 2.193 & 0.212 & 0.788 & 2.114 & \textbf{0.114} & 0.475 & 0.925 & \textbf{0.169} & 0.630 & 1.167 & 0.205 & 0.873 & 1.242 & \textbf{0.070} & 0.366 & 0.720 & \textbf{0.095} & 0.327 & 0.512 & \textbf{0.028} & 0.133 & 0.506 \\
        \hline
        BERT & 0.337 & 1.131 & 2.164 & 0.509 & 1.395 & 2.315 & 0.233 & 0.800 & 2.111 & 0.184 & 0.506 & 0.958 & 0.256 & 0.677 & 1.214 & \textbf{0.175} & 0.771 & 1.213 & 0.107 & 0.374 & 0.707 & 0.129 & 0.346 & 0.516 & 0.040 & 0.115 & 0.458 \\
        $\Delta$-interpolator & 0.287 & 1.124 & 2.176 & 0.391 & 1.297 & 2.183 & 0.217 & 0.780 & 2.093 & 0.125 & 0.498 & 0.938 & 0.193 & 0.650 & 1.145 & 0.242 & 0.650 & 1.369 & 0.083 & 0.375 & 0.757 & 0.124 & 0.355 & 0.565 & 0.049 & 0.152 & 0.616 \\
        $\text{TG}_{\text{complete}}$ & 0.348 & 1.117 & 2.136 & 0.485 & 1.373 & 2.270 & 0.193 & 0.737 & 2.120 & 0.234 & 0.533 & 0.924 & 0.321 & 0.732 & 1.231 & 0.268 & 0.781 & 1.253 & 0.138 & 0.370 & 0.700 & 0.167 & 0.368 & 0.568 & 0.066 & 0.135 & 0.369 \\
        MAE & 0.279 & 0.864 & 1.930 & 0.595 & 1.188 & 2.058 & 0.368 & 0.904 & 2.623 & 0.281 & 0.508 & 0.820 & 0.662 & 0.903 & 1.222 & 0.612 & 0.893 & 1.536 & 0.118 & 0.309 & 0.554 & 0.154 & 0.315 & 0.459 & 0.068 & 0.138 & 0.336 \\
        \hline
        Ours & \textbf{0.205} & \textbf{0.662} & \textbf{1.519} & 0.393 & \textbf{0.823} & \textbf{1.417} & \textbf{0.186} & \textbf{0.588} & \textbf{1.707} & 0.134 & \textbf{0.307} & \textbf{0.583} & 0.212 & \textbf{0.453} & \textbf{0.770} & 0.186 & \textbf{0.431} & \textbf{0.945} & 0.076 & \textbf{0.188} & \textbf{0.400} & 0.103 & \textbf{0.222} & \textbf{0.362} & 0.044 & \textbf{0.091} & \textbf{0.247} \\
        \hline
    \end{tabular}
    }
    \resizebox{\textwidth}{!}{
    \begin{tabular}{|c|c c c|c c c|c c c|c c c|c c c|c c c|c c c|c c c|c c c|}
        \hline
        Category & \multicolumn{9}{c|}{LaFAN1:crawl} & \multicolumn{9}{c|}{LaFAN1:dance} & \multicolumn{9}{c|}{LaFAN1:get up} \\
        \hline
        LERP & \textbf{0.088} & 0.481 & 1.045 & \textbf{0.093} & 0.365 & 0.671 & \textbf{0.082} & 0.512 & 1.541 & 0.141 & 0.683 & 1.287 & \textbf{0.150} & 0.583 & 0.984 & 0.111 & 0.516 & 1.232 & \textbf{0.095} & 0.505 & 1.118 & \textbf{0.104} & 0.402 & 0.704 & \textbf{0.072} & 0.458 & 1.385 \\
        \hline
        BERT & 0.181 & 0.522 & 1.075 & 0.166 & 0.403 & 0.699 & 0.143 & 0.422 & 1.432 & 0.209 & 0.717 & 1.311 & 0.212 & 0.622 & 1.012 & 0.100 & 0.467 & 1.158 & 0.190 & 0.545 & 1.147 & 0.182 & 0.446 & 0.734 & 0.099 & 0.394 & 1.290 \\
        $\Delta$-interpolator & 0.097 & 0.437 & 0.970 & 0.106 & 0.351 & 0.660 & 0.120 & 0.518 & 1.497 & 0.164 & 0.703 & 1.308 & 0.178 & 0.607 & 1.014 & 0.139 & 0.507 & 1.255 & 0.126 & 0.584 & 1.279 & 0.136 & 0.459 & 0.797 & 0.098 & 0.442 & 1.345 \\
        $\text{TG}_{\text{complete}}$ & 0.203 & 0.500 & 0.996 & 0.180 & 0.413 & 0.719 & 0.207 & 0.506 & 1.309 & 0.244 & 0.729 & 1.332 & 0.232 & 0.630 & 1.024 & 0.130 & 0.455 & 1.077 & 0.200 & 0.493 & 0.960 & 0.186 & 0.434 & 0.726 & 0.153 & 0.439 & 1.051 \\
        MAE & 0.244 & 0.497 & 0.909 & 0.243 & 0.408 & 0.633 & 0.321 & 0.501 & 1.141 & 0.225 & 0.640 & 1.069 & 0.251 & 0.569 & 0.871 & 0.189 & 0.468 & 0.961 & 0.263 & 0.513 & 0.913 & 0.269 & 0.446 & 0.668 & 0.276 & 0.443 & 0.891 \\
        \hline
        Ours & 0.115 & \textbf{0.343} & \textbf{0.681} & 0.136 & \textbf{0.312} & \textbf{0.530} & 0.127 & \textbf{0.324} & \textbf{0.858} & \textbf{0.128} & \textbf{0.506} & \textbf{0.917} & 0.168 & \textbf{0.494} & \textbf{0.777} & \textbf{0.087} & \textbf{0.357} & \textbf{0.829} & 0.135 & \textbf{0.340} & \textbf{0.645} & 0.150 & \textbf{0.329} & \textbf{0.543} & 0.105 & \textbf{0.265} & \textbf{0.664} \\
        \hline
    \end{tabular}
    }
    \resizebox{\textwidth}{!}{
    \begin{tabular}{|c|c c c|c c c|c c c|c c c|c c c|c c c|c c c|c c c|c c c|}
        \hline
        Category & \multicolumn{9}{c|}{LaFAN1:jump/hop} & \multicolumn{9}{c|}{LaFAN1:obstacles} & \multicolumn{9}{c|}{LaFAN1:walk} \\
        \hline
        LERP & 0.178 & 0.837 & 1.327 & \textbf{0.146} & 0.544 & 0.779 & 0.073 & 0.304 & 0.642 & 0.153 & 0.796 & 1.616 & \textbf{0.126} & 0.497 & 0.818 & 0.049 & 0.256 & 0.757 & 0.124 & 0.669 & 1.451 & \textbf{0.117} & 0.467 & 0.793 & \textbf{0.052} & 0.265 & 0.950 \\
        \hline
        BERT & 0.277 & 0.886 & 1.331 & 0.222 & 0.584 & 0.803 & 0.092 & 0.280 & 0.602 & 0.227 & 0.830 & 1.629 & 0.183 & 0.529 & 0.834 & 0.062 & 0.230 & 0.696 & 0.182 & 0.682 & 1.442 & 0.169 & 0.492 & 0.801 & 0.059 & 0.228 & 0.882 \\
        $\Delta$-interpolator & 0.173 & 0.777 & 1.331 & 0.153 & 0.520 & 0.811 & 0.094 & 0.301 & 0.727 & 0.150 & 0.732 & 1.484 & 0.134 & 0.486 & 0.805 & 0.070 & 0.271 & 0.879 & 0.134 & 0.647 & 1.389 & 0.135 & 0.471 & 0.799 & 0.067 & 0.242 & 0.872 \\
        $\text{TG}_{\text{complete}}$ & 0.299 & 0.854 & 1.391 & 0.244 & 0.608 & 0.923 & 0.136 & 0.401 & 0.628 & 0.247 & 0.640 & 1.246 & 0.205 & 0.475 & 0.770 & 0.108 & 0.290 & 0.643 & 0.217 & 0.570 & 1.148 & 0.188 & 0.446 & 0.742 & 0.092 & 0.248 & 0.566 \\
        MAE & 0.275 & 0.737 & 1.123 & 0.262 & 0.536 & 0.757 & 0.111 & 0.299 & 0.585 & 0.263 & 0.574 & 1.077 & 0.250 & 0.426 & 0.643 & 0.118 & 0.228 & 0.466 & 0.219 & 0.525 & 0.912 & 0.224 & 0.414 & 0.597 & 0.135 & 0.261 & 0.578 \\
        \hline
        Ours & \textbf{0.151} & \textbf{0.557} & \textbf{0.940} & 0.163 & \textbf{0.455} & \textbf{0.677} & \textbf{0.052} & \textbf{0.216} & \textbf{0.450} & \textbf{0.130} & \textbf{0.366} & \textbf{0.789} & 0.138 & \textbf{0.307} & \textbf{0.510} & \textbf{0.048} & \textbf{0.129} & \textbf{0.335} & \textbf{0.111} & \textbf{0.322} & \textbf{0.639} & 0.128 & \textbf{0.284} & \textbf{0.456} & 0.054 & \textbf{0.139} & \textbf{0.344} \\
        \hline
    \end{tabular}
    }
    \caption{Comparison of average L2P, L2Q, and NPSS performance on motion samples from various categories, each with 121 frames in length. Lower values indicate more accurate predictions. The top performer of each test is highlighted in \textbf{bold}.}
    \label{tab:perf}
\end{table*}

\subsection{Datasets and metrics}

We benchmark our method in the motion interpolation task against both the state-of-the-art RNN-based network \cite{harvey2020robust} and BERT-based network \cite{duan2021single} in motion transition generation. To evaluate the effectiveness of each model, they are to complete the following motion interpolation task:

\begin{itemize}
    \item \textbf{Input}: The model is provided with the keyframes $K$ of an $N$-frame ground truth motion. While keyframes can be defined for any combination of frames, we place each keyframe evenly every 5, 15, or 30 frames for consistency, starting with the first frame, e.g., $K = \{x_0, x_5, x_{10}, ..., x_{|K|}\}$ for 5-frame intervals.
    \item \textbf{Expected output}: Each model is to output an $N$-frame motion $\hat{Y} = \{\hat{y}_0, \hat{y}_1, ..., \hat{y}_{N-1}\}$ given the keyframes $K$. This output is compared with the ground truth with the L2P and L2Q metrics used in state-of-the-art comparisons \cite{harvey2020robust,duan2021single}. These metrics measure the average $\ell_2$ errors of all positional and rotational attributes respectively for each pose. In addition, we apply the Normalised Power Spectrum Similarity (NPSS) \cite{gopalakrishnan2019neural} metric to measure visual similarities between the estimated and actual motion outputs.
\end{itemize}

We source our motions for both training and evaluation from the Ubisoft La Forge Animation (LaFAN1) dataset \cite{harvey2020robust} and the Carnegie-Mellon University Motion Capture (CMU Mocap) dataset {\footnote{Dataset available at: \url{http://mocap.cs.cmu.edu/}}}. The CMU Mocap motions are resampled from their original 120 frames per second (FPS) down to 30 FPS considering the computational cost, and to match the frame rate of the LaFAN1 dataset. While LaFAN1 focuses largely on motions with visibly dynamic details such as locomotion and sports actions, CMU Mocap provides a larger variety of motions, many of which exhibit more minute details and ambiguous motion trajectories. From a data-level perspective, this suggests that root position accuracy is more important in the LaFAN1 dataset compared to the CMU Mocap dataset. We employ both datasets to demonstrate our method's ability to adapt with different levels of motion dynamics.

For our model, the positional data of each motion sample is recentred around the $XYZ$ means of the keyframed root positions. Given that unit quaternion values are restricted to a range of $[-1, 1]$, while positional values can be of any scalar value, we rescale positional data such that $L_{\text{root}} \approx L_{\text{quat}}$ with initial model weights. During training, we randomly select between $[\lfloor \frac{|Y|}{24} \rfloor, \lfloor \frac{|Y|}{4} \rfloor]$ keyframes for each sampled motion $Y$, with the first and last frames being stipulated as keyframes. Each motion sample batch has a random length of $|Y| \in [72, 144]$. 

\subsection{Implementation details}

We implement our method with 8 layers for each transformer, with a token embedding size of $d = 512$, and an FFN size of $4 \times d$. Each multi-head attention layer is split between 8 attention heads. Since our method relies on coordination between Stage-I and Stage-II, the training stability of deeper models benefits greatly from larger batch sizes. For our 8-layer setting, we found that a batch size of 64 motions per epoch is sufficient for convergence.

We train our model using the Adam optimiser \cite{kingma2014adam} for 50,000 epochs with a scheduled learning rate. Specifically, we employ both a warm-up and decay strategy for our learning rate using the following strategy \cite{vaswani2017attention}:
\begin{equation}
    lr(e) = 4\text{e-}4 \times \text{min}(e^{-\frac{1}{2}}, e \times 1000^{-\frac{3}{2}}),
\end{equation}
where $e$ represents the number of current training epoch. In addition, we linearly scale $\alpha_g \in [0, 1]$ for 1,000 epochs after warm-up, in order to avoid conflicting gradients caused by ambiguous quaternion polarity when backpropagating through FK functions, i.e. $FK(Q) = FK(-Q)$ for any set of joint rotation quaternions $Q$. We set $\alpha_l = 1$ throughout the training process.

\begin{table}[]
    \centering
    \setlength\tabcolsep{2pt}
    \resizebox{0.49\textwidth}{!}{
    \begin{tabular}{|c|c|c|c|c|c c c|}
        \hline
        & \textbf{$\Phi^{\text{syn}}$ Key \& Intermediate Input} & \textbf{Layers} & \textbf{Norm} & \textbf{PE mode} & \textbf{L2P} & \textbf{L2Q} & \textbf{NPSS} \\
        \hline
        (a) & N/A - No $\Phi^{\text{syn}}$ & 8 & Seq-RC & concat & 0.665 & 0.699 & 0.640 \\
        \hline
        (b) & $\Phi^{\text{key}}$ \& $\Phi^{\text{imd}}$ & 8 & LayerNorm & concat & 0.539 & 0.620 & 0.581 \\
        \hline
        (c) & $\Phi^{\text{key}}$ \& $\Phi^{\text{imd}}$ & 8 & Seq-RC & additive & 0.587 & 0.661 & 0.565 \\
        \hline
        \multirow{8}{*}{(d)} & \multirow{4}{*}{\text{keyframe embeddings} \& $\Phi^{\text{imd}}$} & 2 & \multirow{4}{*}{Seq-RC} & \multirow{4}{*}{concat} & 0.684 & 0.711 & 0.675 \\
        & & 4 & & & 0.610 & 0.642 & 0.532 \\
        & & 8 & & & \multicolumn{3}{c|}{\textit{Divergence}} \\
        & & 12 & & & \multicolumn{3}{c|}{\textit{Divergence}} \\
        \cline{2-8}
        & \multirow{4}{*}{$\Phi^{\text{key}}$  \& $\Phi^{\text{imd}}$} & 2 & \multirow{4}{*}{Seq-RC} & \multirow{4}{*}{concat} & 0.652 & 0.691 & 0.609 \\
        & & 4 & & & 0.534 & 0.608 & 0.501 \\
        & & 8 & & & 0.438 & 0.530 & 0.384 \\
        & & 12 & & & \textbf{0.416} & \textbf{0.515} & \textbf{0.380} \\
        \hline
    \end{tabular}
    }
    \caption{Ablation study of our architectural features: (a) Stage-III manifold refinement (b) Seq-RC, (c) concatenated PE, (d) bridging $\Phi^{\text{key}}$ into $\Phi^{\text{syn}}$, convergability \& depth comparison.}
    \label{tab:layers}
\end{table}

\begin{figure}[]
     \centering
     \includegraphics[width=\linewidth]{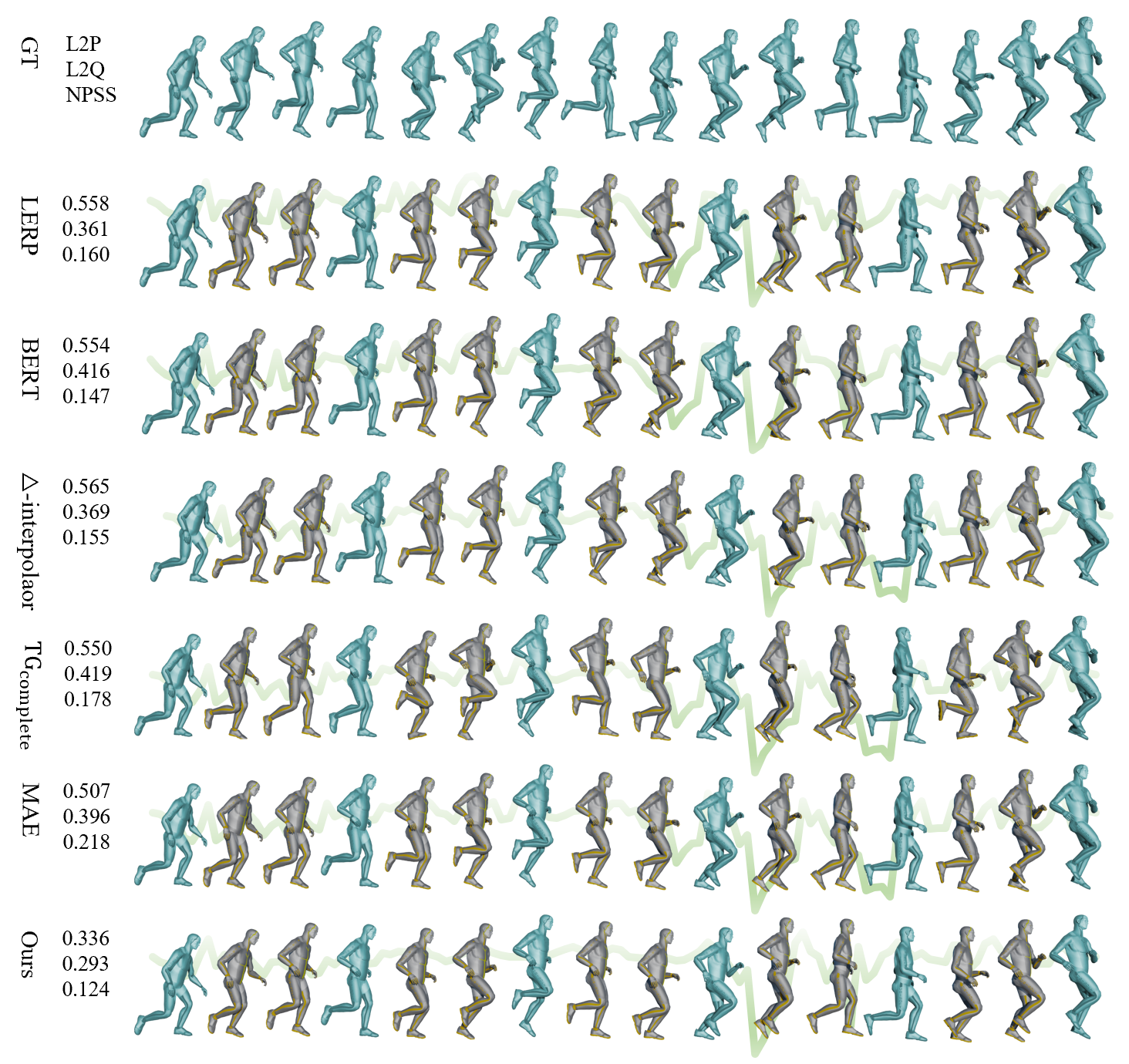}
     \caption{An example of motion interpolation by each of the tested methods, 
     compared with the ground truth motion (first row). The green curves indicate the offsets regarding the $\ell_1$ distance between the interpolation and ground truth manifolds. Less turbulent offsets indicate more visually similar motion predictions. 
     }
     \label{fig:example_hop}
\end{figure}

\subsection{Comparison to the state-of-the-art}

We benchmark the performance of our method against state-of-the-art models by evaluating their L2P, L2Q, and NPSS metrics with testing datasets. 
Table \ref{tab:perf} compares the performance of our architecture against the BERT-based motion in-painting transformer \cite{duan2021single}, the encoder-decoder-based $\Delta$-interpolator \cite{oreshkin2022motion}, the RNN-based approach $\text{TG}_{\text{complete}}$ \cite{harvey2020robust}, and the masked auto-encoder (MAE) architecture \cite{he2022masked}.

The quantitative performance of our model is greatly improved over all existing methods, as well as LERP, in a large majority of keyframing scenarios. 
The performance improvement of our method compared to LERP increases with the length of keyframe intervals, as using learning based methods provides the opportunity to reconstruct non-linear motion details. 
Note that for a short keyframe interval, a linear estimation ($f(x+\Delta x)=f(x)+\hat{f}'(x)\Delta x + o(\Delta x)$) of a continuous motion (function) can be relatively accurate, which explains why similar performance is found between LERP and our method for the 5-frame interval setting. However, other existing methods are significantly worse than LERP. 
Our model notably outperforms both $\text{TG}_{\text{complete}}$ and MAE with its single token mask in every scenario. Thus, a clear motion interpolation improvement can be observed from our decoupling strategy with motion manifold technique, compared to the RNN model.

The BERT-based method \cite{duan2021single} exhibits a clear disadvantage in its performance due to its reliance on LERP-based input mask tokens. 
By deriving the mask token embeddings from a sub-optimal estimation, self-attention mechanisms tend to converge towards reproducing the input token rather than composing more realistic poses, as it is close to a trivial local minimum to learn. Consequently, such models never learn to fully consider the keyframe tokens as their main source of information. 
Figure \ref{fig:example_hop} highlights the near-identical latent manifolds of LERP output and BERT-based evaluation.
We observe a similar behaviour with the $\Delta$-interpolator model, where LERP-based transformations are applied as a post-processing step \cite{oreshkin2022motion}. While its $\Delta$-mode strategy allows the model to perform marginal improvements over LERP more frequently, it is still heavily reliant on the performance of LERP, which does not bode well with longer keyframe intervals. To this end, we can deduce that the realistic interpolation can be difficult with LERP-reliant solutions.
Conversely, our manifold learning approach fully considers the continuous joint positions and rotations of the input keyframes, and is able to converge upon a significantly more optimal solution.

\begin{table}[]
    \centering
    \small
    \resizebox{0.35\textwidth}{!}{
    \begin{tabular}{|p{2.5cm}|c|c c c|}
        \hline
        $|Y|$ & \# Params & 31 & 61 & 121 \\
        \hline
        LERP & - & 0.0330 & 0.0360 & 0.0370 \\
        $\text{TG}_{\text{complete}}$ & 15.6M & 0.5001 & 1.0272 & 1.9774 \\
        BERT & 29.3M & 0.0541 & 0.0570 & 0.0596 \\
        MAE & 54.9M & 0.0755 & 0.0793 & 0.0820 \\
        Ours & 83.2M & 0.0793 & 0.0830 & 0.0850 \\
        \hline
    \end{tabular}
    }
    \caption{Inference time in seconds and parameter count for different motion lengths $|Y|$. Keyframes of each evaluation were evenly placed every 15 frames, starting from the first frame.}
    \label{tab:elapsed}
\end{table}

\begin{figure}[t]
    \centering
    \includegraphics[width=0.8\linewidth]{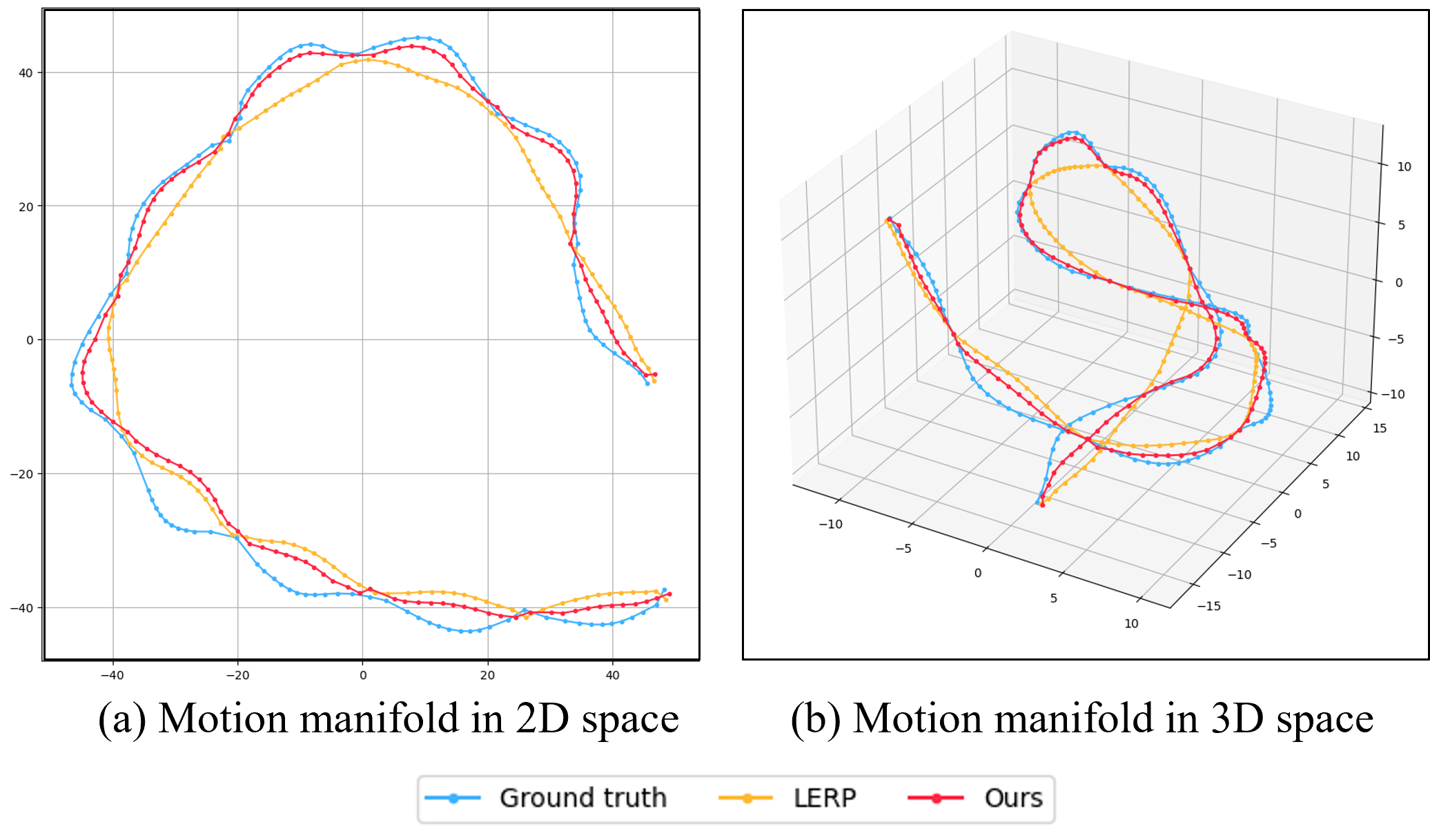}
    \caption{Sample motion manifolds obtained by t-SNE.}
    \label{fig:manifold}
\end{figure}

\begin{figure}[t]
    \centering
    \includegraphics[width=\linewidth]{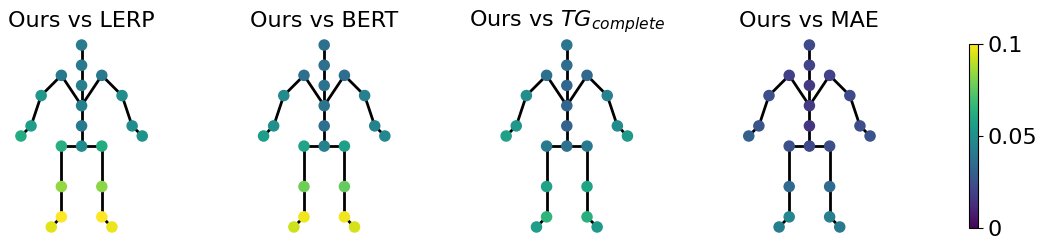}
    \caption{Performance improvement of our architecture by joints in L2P + L2Q, compared to existing methods.}
    \label{fig:joints}
\end{figure}

Table \ref{tab:layers} documents an ablation study for each our architecture's contributions. Major improvements to the architecture's L2P, L2Q, and NPSS performance can be observed for the inclusion of manifold self-attention in Stage III, the replacement of LayerNorm with our sequence-level re-centering normalisation scheme, and concatenation over addition of PE. It should be noted that the Stages I and II only variant of our model (i.e., (a) in Table \ref{tab:layers}) is structurally identical to the $\Delta$-interpolator model with $\Delta$-mode disabled \cite{oreshkin2022motion}. In addition, we demonstrate the use of $\Phi_{\text{key}}$ token representations over separately trained keyframe embeddings in Stage-III, which leads to improved convergibility of deeper architecture settings. We further demonstrate the importance of such deeper settings, which provide a significant boost to our model's evaluation accuracy.

Figure \ref{fig:manifold} visualises the latent motion manifolds in 2D and 3D spaces for our method, LERP and ground truth using t-SNE analysis. The manifold of our method is obtained from the inputs of Stage-III, and LERP and ground truth are from the pose data. It can be observed that the lower-dimensional curves (i.e., manifolds) represent the motion of higher-dimensional in a smooth manner, and ours is very close to the one associated with the ground truth, compared with LERP. 
This indicates the superiority of our method to derive a high-quality latent motion space. Figure \ref{fig:example_hop}
illustrates an example of hopping motion interpolated by different methods with quantitative metrics. The ground truth data is reduced to a motion manifold with t-SNE. The offset of the manifold from each interpolation method compared to the ground truth one is obtained by $\ell_1$ distance for visualization. Particularly, the offset values are enlarged for observation purpose. Our method gains the best motion manifold with the least offset from the ground truth manifold. 

\begin{table}[]
    \centering
    \resizebox{0.475\textwidth}{!}{
    \begin{tabular}{|c|c c c|c c c|c c c|}
        \multicolumn{1}{c}{} & \multicolumn{3}{c}{\textbf{L2P}} & \multicolumn{3}{c}{\textbf{L2Q}} & \multicolumn{3}{c}{\textbf{NPSS}} \\
        \hline
        Interval & 5 & 15 & 30 & 5 & 15 & 30 & 5 & 15 & 30 \\
        \hline
        LERP & 0.35 & 1.28 & 2.46 & 0.22 & 0.66 & 1.17 & 0.0021 & 0.0430 & 0.2663 \\
        $\text{TG}_{\text{complete}}$ & 0.22 & 0.64 & 1.25 & 0.17 & 0.45 & 0.68 & 0.0019 & 0.0247 & 0.1298 \\
        BERT & 0.22 & 0.60 & 1.14 & 0.15 & 0.38 & 0.60 & 0.0016 & 0.0251 & 0.1270 \\
        $\Delta$-interpolator & 0.16 & 0.53 & 1.05 & 0.12 & 0.33 & 0.59 & 0.0015 & 0.0238 & 0.1272 \\
        \hline
        Ours & 0.30 & 0.71 & 1.26 & 0.21 & 0.40 & 0.63 & 0.0019 & 0.0284 & 0.1393 \\
        \hline
    \end{tabular}
    }
    \caption{Motion completion performance of our method, based on the Harvey et al. (2020) \cite{harvey2020robust} setup.}
    \label{tab:inbetween}
\end{table}

Figure \ref{fig:joints} dissects the L2P and L2Q improvements of our method into individual joints. We can clearly observe that the main improvements of our method over LERP exist within the global positions and rotations of foot joints, whereas improvements are more widely spread compared to the MAE and RNN-based methods. On average, our approach brings a L2P and L2Q benefit to all joints.

\subsection{Inference latency}

The inference time of different motion interpolation methods was evaluated in our experiments, as the visual latency of keyframe adjustments is important for efficient animation work, as well as real-time applications. 
We implemented these methods with PyTorch on an AMD Ryzen 9 3950X processor and an NVIDIA GeForce RTX 3090 GPU. Table \ref{tab:elapsed} shows that the parallelism provided by the transformer is highly beneficial to our method when interpolating complete motion sequences. Our method shows a similar order of time complexity to the LERP method, being consistently around $2.5\times$ LERP inference time, and significantly faster than the RNN-based approach. 
In addition, our method is stable in terms of the inference time for different sequence lengths, whilst the RNN-based approach shows a significant increasing latency from 31-frame sequence to 121-frame sequence. 

\subsection{Extension to motion completion}

Though our model is designed for sparse keyframe interpolation, we can additionally perform motion completion as it can be defined as a specifically constructed keyframe set. Table \ref{tab:inbetween} compares the performance of our model against linear interpolation and the state-of-the-art models for motion completion. With the benefit of the motion context, our model outperforms LERP, but not to the efficacy of the RNN-based \cite{harvey2020robust} and transformer-based \cite{duan2021single, oreshkin2022motion} models.

\subsection{Limitations and future work}

One limitation of our approach is that its maximum motion length is limited by the length of its training samples. Unlike most transformer-based solutions that can trivially employ relative positional encodings \cite{dai2019transformer, shaw2018self}, our method relies on continuous positional vectors, such as sinusoidal encodings, for its manifold representations, and thus cannot employ the same model to accept inputs of arbitrary length. Further research for a compatible relative position representation would allow our approach to be seamlessly applied in keyframed animation workflows for longer sequences.

\section{Conclusion}

This paper presents a three-stage transformer-based motion interpolation method. We begin by producing learned motion keyframe context tokens in Stage-I. With context-guided attention, Stage-II generates embeddings for intermediate tokens by inferencing an implicitly constrained latent motion manifold with the guidance of the keyframe context tokens. Stage-III takes both the keyframe tokens and intermediate tokens to compose the interpolated motion sequence. In addition, we introduce a novel sequence-level re-centreing technique to address the feature biases that are more prevalent in sequences of continuous attributes. We demonstrate that the superior interpolation accuracy of our approach compared with existing RNN and masked transformer methods. 
As our architecture is designed for any masked sequence-to-sequence task with continuous attributes, we believe that our architecture's applications extend beyond motion interpolation.

\section*{Acknowledgment}
This research was in part 
supported by Australian Research Council (ARC) grant \#DP210102674.

{\small
\bibliographystyle{ieee_fullname}
\bibliography{egbib}
}

\end{document}